%% file: main.tex
\documentclass{article}

\usepackage{microtype}
\usepackage{subfigure}
\usepackage{adjustbox}
\usepackage[accepted]{icml2025}

\usepackage{amsmath}
\usepackage{amssymb}
\usepackage{mathtools}
\usepackage{amsthm}
\usepackage{xspace}
\usepackage{amsmath}

\allowdisplaybreaks

\usepackage{hyperref}
\usepackage[capitalize,noabbrev]{cleveref}

\usepackage{natbib}
\PassOptionsToPackage{table}{xcolor}

\usepackage{url}
\usepackage{booktabs}
\usepackage{graphicx}
\usepackage{algorithmic}
\usepackage{tikz}
\usetikzlibrary{tikzmark,calc}
\usepackage{multirow}
\usepackage{xcolor}
\usepackage{colortbl}
\usepackage{pifont} 
\usepackage{makecell}
\definecolor{lightblue}{rgb}{0.1, 0.1, 0.9} 
\usepackage{fancyhdr}
\usepackage[normalem]{ulem}
\usepackage{adjustbox}

\theoremstyle{plain}
\theoremstyle{definition}

\theoremstyle{remark}

\usepackage[textsize=tiny]{todonotes}

\icmltitlerunning{SageAttention2++: A More Efficient Implementation of SageAttention2} % Submission and Formatting Instructions for ICML 2024

\newcommand{\ourf}{\texttt{SageAttn2(4+8)}\xspace}
\newcommand{\oure}{\texttt{SageAttn2(8+8)}\xspace}
\newcommand{\oura}{\texttt{SageAttention2++}\xspace}
\newcommand{\ourfa}{\texttt{SageAttn2++(4+8)}\xspace}
\newcommand{\ourea}{\texttt{SageAttn2++(8+8)}\xspace}

\NewDocumentCommand{\jintao}{ mO{} }{\textcolor{blue}{\textsuperscript{\textit{JT}}\textsf{\small[#1]}}}

\newcommand{\cogvideo}{\texttt{CogvideoX}\xspace}
\newcommand{\hyvideo}{\texttt{HunyuanVideo}\xspace}

\newcommand{\wan}{\texttt{Wan}\xspace}

\newcommand{\llamal}{\texttt{Llama3.1}\xspace}

\newcommand{\flux}{\texttt{Flux}\xspace}
\newcommand{\sd}{\texttt{Stable-Diffusion3.5}\xspace}

\definecolor{deepgreen}{rgb}{0.0, 0.5, 0.0}  
\definecolor{deepred}{rgb}{0.6, 0.0, 0.0}

\begin{document}

\twocolumn[
\icmltitle{SageAttention2++: A More Efficient Implementation of SageAttention2}



\icmlsetsymbol{equal}{*}

\begin{icmlauthorlist}
\icmlauthor{Jintao Zhang}{yyy}
\icmlauthor{Xiaoming Xu}{yyy}
\icmlauthor{Jia Wei}{yyy}
\icmlauthor{Haofeng Huang}{yyy}
\icmlauthor{Pengle Zhang}{yyy}
\icmlauthor{Chendong Xiang}{yyy} \\
\icmlauthor{Jun Zhu}{yyy}
\icmlauthor{Jianfei Chen}{yyy}
\end{icmlauthorlist}

\icmlaffiliation{yyy}{Department of Computer Science, Tsinghua University}
% \icmlaffiliation{comp}{Institute for Interdisciplinary Information Sciences, Tsinghua University}

% \icmlcorrespondingauthor{Jianfei Chen}{jianfeic@tsinghua.edu.cn} %{first1.last1@xxx.edu}

\icmlkeywords{Machine Learning, ICML}

\vskip 0.3in
]



\printAffiliationsAndNotice{}  % leave blank if no need to mention equal contribution

\begin{abstract}
The efficiency of attention is critical because its time complexity grows quadratically with sequence length. SageAttention2 addresses this by using quantization to speed up matrix multiplications (Matmul) in attention. To further accelerate SageAttention2, we propose to utilize the faster instruction of FP8 Matmul accumulated in FP16. The instruction is 2$\times$ faster than the FP8 Matmul used in SageAttention2.
Our experiments show that \oura achieves a \textbf{3.9}$\times$ speedup over FlashAttention while maintaining the same attention accuracy as SageAttention2. This means \oura effectively accelerates various models, including those for language, image, and video generation, with negligible end-to-end metrics loss. The code will be available at \url{https://github.com/thu-ml/SageAttention}.
\end{abstract}

\input{src/Introduction}

\input{src/Preliminary}

\input{src/Method}

\input{src/Experiment}

\input{src/Conclusion}

\newpage
\bibliography{main}
\bibliographystyle{icml2025}
\newpage

\appendix
\onecolumn

\input{src/Appendix}

\end{document}

%% file: src/Introduction.tex
\section{Introduction}
The quadratic time complexity of attention necessitates efficient implementations for real-world applications with long sequences~\cite{jiang2024minference}. Current approaches to reduce attention's computational demands fall into three main categories: (1) \textit{linear attention} methods~\cite{wang2020linformer,choromanski2020rethinking,yu2022metaformer,katharopoulos2020transformers,qin2024lightning,yang2024gated} that achieve $O(N)$ complexity, and (2) \textit{sparse attention} techniques~\cite{liu2021swin,chu2021twins,liuniformer,xiao2023efficient,xiao2024infllm,chen2023longlora,jiang2024minference,venkataramanan2023skip,gao2024seerattention,moaattention,zhang2025spargeattn,xi2025sparse,zhang2025spargeattn_wksp} that process only relevant context portions. While effective, these methods often exhibit limited generality across models and tasks.  
(3) An alternative direction focuses on hardware-optimized attention implementations that maintain full sequence computation while achieving superior speed and accuracy. Notable examples include FlashAttention~\cite{dao2022flashattention}, FlashAttention2~\cite{dao2023flashattention,shah2024flashattention}, xFormers~\cite{xFormers2022}, and SageAttentions~\cite{2024sageattention,2024sageattention,zhang2025sageattention3,zhang2025sageattention2_wksp} which demonstrate strong performance across diverse applications.

\noindent\textbf{{Motivation, problem, and our approach}}. For the second matrix multiplication (Matmul) $PV$ in attention, SageAttention2 accelerates it by quantizing to FP8 and using the \texttt{mma.f32.f8.f8.f32} instruction. However, \texttt{mma.f32.f8.f8.f32} employs an FP32 accumulator and is only 2$\times$ faster than FP16.  We find that the \texttt{mma.f16.f8.f8.f16} instruction (using FP16 accumulator for FP8 Matmul) achieves 4$\times$ speedup over FP16~\cite{nvidia2022ada}.  Therefore, we aim to accelerate SageAttention2 by using the faster instruction. However, directly using the faster instruction will lead to the values of $PV$ exceeding the representable range of FP16. To address the problem, we propose to narrow the quantization range of $P$ and $V$ to satisfy the accumulator range in FP16.

\noindent\textbf{{Performance.}}  For efficiency, \oura delivers a 3.9$\times$ speedup over FlashAttention. In terms of accuracy, \oura matches SageAttention2's performance.
We conduct comprehensive evaluations on state-of-the-art models for text, image, and video generation. The results demonstrate that \oura provides plug-and-play acceleration with negligible end-to-end metrics loss across diverse models.

%% file: src/Preliminary.tex
\section{Preliminary} \label{sec:preliminary}

\begin{table}[htb!]
    \centering
    \caption{Speedup compared to matrix multiplication in FP16 with an FP32 accumulator.}
    \label{tab:speedup_of_tensor_core}
    \setlength\tabcolsep{5pt}
    \scalebox{0.96}{
    \begin{tabular}{c|c|c|c}
        \toprule
        \textbf{GPU} & \textbf{MM Input} & \textbf{MM Accumulator} & \textbf{Speedup} \\ \hline
        \multirow{3}{*}{\makecell[c]{RTX4090, \\ RTX5090}}  
        % & INT8	 & INT32  &  2x \\
          & FP16	 & FP32  &  \textbf{1x} \\
          & FP8	 & FP32  &  \textbf{2x} \\
          & FP8	 & \textbf{FP16}  &  \textbf{4x }\\
        \bottomrule                  
    \end{tabular}
    }
\end{table}

\begin{table}[htb!]
      \centering
      \caption{Average attention accuracy across all attention layers of CogvideoX.}
      \label{tab:accuracy_of_different_PrVr}
      \setlength\tabcolsep{8.1pt}
      \scalebox{0.97}{
      \begin{tabular}{c|c|c|c|c}
          \toprule
          \textbf{Method} & \textbf{$P_r$} & \textbf{$V_r$} & \textbf{Cossim$\uparrow$} & \textbf{L1$\downarrow$} \\ \hline
          SageAttn2  & 448	 & 448  &  99.97\% & 0.01862 \\
          SageAttn2++  & 448	 & 2.25  &  99.97\% & 0.01863 \\
          SageAttn2++  & 224	 & 4.5  &  99.97\% & \textbf{0.01862} \\
          SageAttn2++  & 112	 & 9  &  99.97\% & 0.01863 \\
          \bottomrule                  
      \end{tabular}
      }
  \end{table}

\subsection{SageAttention2}

SageAttention2~\cite{zhang2024sageattention2} is a quantization~\cite{zhang2025int8train,hu2025quant} method based on FlashAttention~\cite{dao2022flashattention}. FlashAttention tiles $Q, K, P, V$ into blocks ($\{Q_i\}, \{K_i\}, \{\widetilde P_i\}, \{V_i\}$) and uses online softmax~\cite{milakov2018online} to compute attention progressively. For simplicity, we omit the subscripts ($\{Q_i\}, \{K_i\}, \{\widetilde P_i\}, \{V_i\}$) in the following content and use $Q, K, \widetilde P, V$ to represent the tiled blocks.
SageAttention2 quantizes $Q, K$ to INT4/INT8 with per-block granularity, $\widetilde P$ to FP8 in E4M3 with per-block granularity, and quantizes $V$ to FP8 in E4M3 with per-channel granularity. This means each $Q, K, \widetilde P$ has a separate scale factor: $\delta_{Q} = \max(|{Q|})/127, \delta_{K} = \max(|{K|})/127, \delta_{P} = \max(|{\widetilde P|})/448$, and each channel of $V$ has a separate scalar scale: $\delta_{V} = \mathrm{colmax}(|{V|})/448$. 
By doing so, SageAttention2 accelerates matrix multiplications in attention through low-bit Tensor Core operations. For example, $\hat P = \lceil \widetilde P / \delta_P \rfloor$, $\hat V = \lceil V / \delta_v \rfloor$. Then, $PV = \hat P \hat V * \delta_P * \delta_V$.

\subsection{Data Type of Accumulator for Matmul}

In some GPUs, the speed of Matmul instructions depends on the accumulator data type. For instance, \texttt{mma.f32.f8.f8.f32} uses FP32 accumulator for FP8 Matmul and is only 2$\times$ faster than FP16. The instruction using FP16 accumulator for FP8 Matmul (\texttt{mma.f16.f8.f8.f16}) is 4$\times$ faster than FP16. Table~\ref{tab:speedup_of_tensor_core} summarizes the speedup of Matmul instructions with different accumulators.

%% file: src/Method.tex
\section{SageAttention2++}
In this section, we introduce \oura. The workflow of \oura is based on SageAttention2, also using the smoothing of $Q$ and $K$, INT4/INT8 quantization for $QK^\top$ Matmul and FP8 quantization for $PV$ Matmul. The main difference is that for $PV$, \oura uses the faster instruction (\texttt{mma.f16.f8.f8.f16}), which employs an FP16 accumulator for the FP8 Matmul. To ensure the results of FP8 Matmul remain within FP16's representable range, we adjust the scale factor of the FP8 quantization.

\subsection{Narrowing the FP8 Quantization Range}

The specific MMA~\cite{nvidia2024ptx} instruction used for the MatMul between $P$ and $V$ is \texttt{mma.m16n8k32}. If we quantize $P$ and $V$ to FP8 in E4M3 (-448$\sim$448) as in SageAttention2, the results may exceed FP16's representable range (-65504$\sim$65504). This occurs because 32 product values $pv$ are accumulated in FP16, where $p$ and $v$ come from quantized $\hat P$ and $\hat V$ (derived from $\widetilde P$ and $V$). To ensure the accumulated results stay within FP16's range:
\begin{align}
    |32 \times pv| \le 65504    
\end{align}
For instance, choosing $|p| \le 224$ and $|v| \le 9$ satisfies this condition. We therefore narrow the quantization ranges of $P$ and $V$ by adjusting their scale factors:
\begin{align}
    \delta_P = |\max(\widetilde P)| / P_r,~~~ \delta_V = |\max(V)| / V_r
\end{align}
where we constrain $P_r \times V_r \le 2047$ (since $65504/32 = 2047$).

\subsection{Delayed FP32 Buffering}
The transformation of accumulated values from \texttt{mma.m16n8k32} (in FP16) to FP32 incurs overhead because it needs additional data type conversion PTX instructions~\cite{nvidia2024ptx} to execute. To reduce this overhead, we accumulate two consecutive \texttt{mma.m16n8k32} results in FP16 before performing FP32 conversion, effectively halving the transformation overhead. Maintaining the FP16 representable range requires:
\begin{align}
P_r \times V_r \le 2047/2
\end{align}

\textbf{Choice of $P_r$ and $V_r$}. Table~\ref{tab:accuracy_of_different_PrVr} shows attention accuracy for feasible $(P_r, V_r)$ pairs. The results demonstrate that narrowing quantization ranges introduces negligible error. We select $P_r = 224$ and $V_r = 4.5$ for optimal performance.

%% file: src/Experiment.tex
\begin{figure*}[h!]
    \centering
    \includegraphics[width=1\textwidth]{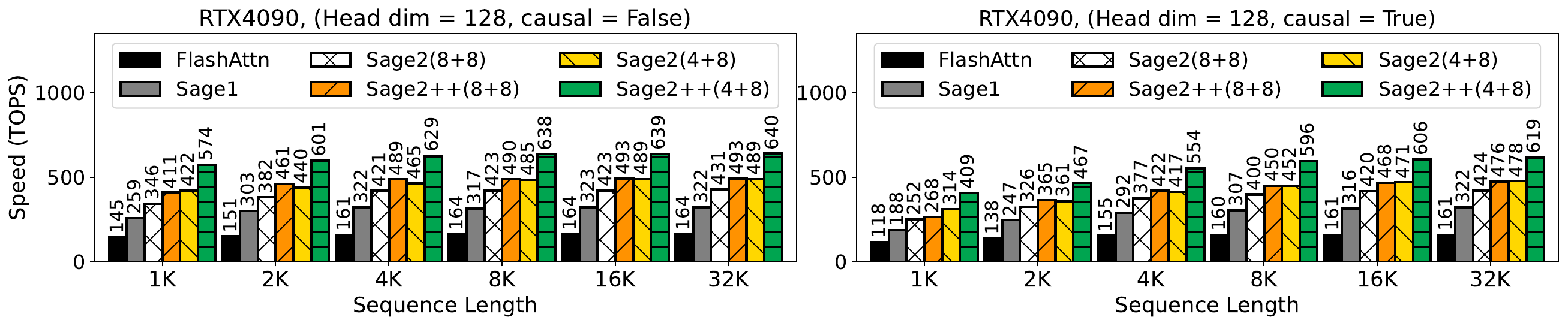}
    \vspace{-1.15em}
    \caption{Speed comparison between \oura and baselines (RTX4090, headdim=128).}
    \vspace{-.5em}
    \label{fig:kf_h128_baseline_RTX4090}
\end{figure*}

\begin{figure*}[h!]
    \centering
    \includegraphics[width=1\textwidth]{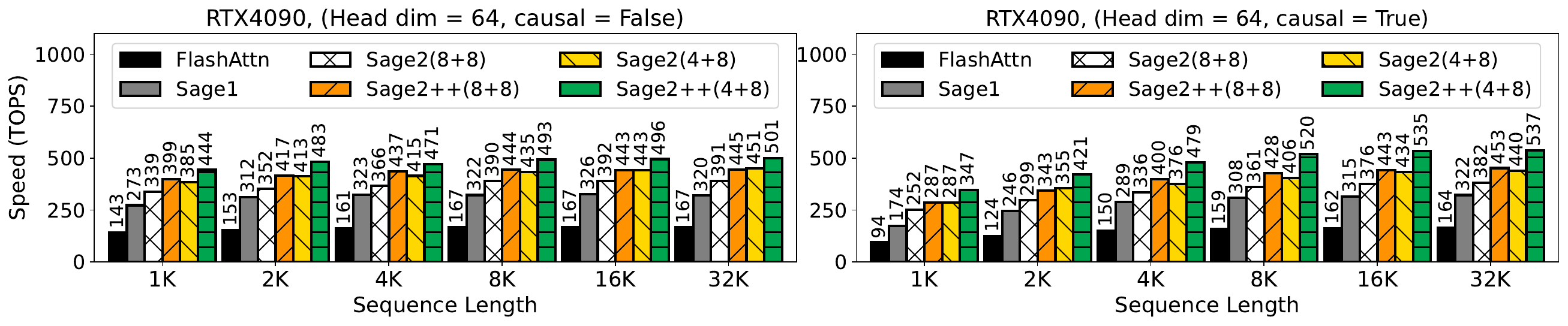}
    \vspace{-1.15em}
    \caption{Speed comparison between \oura and baselines (RTX4090, headdim=64).}
    \vspace{-.5em}
    \label{fig:kf_h64_baseline_RTX4090}
\end{figure*}

\begin{figure*}[h!]
    \centering
    \includegraphics[width=1\textwidth]{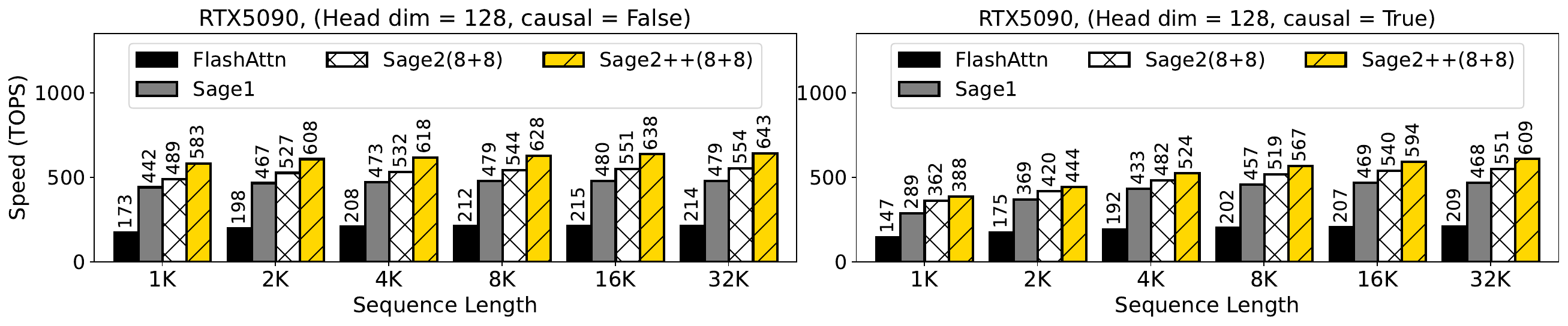}
    \vspace{-1.15em}
    \caption{Speed comparison between \oura and baselines (RTX5090, headdim=128).}
    \vspace{-.5em}
    \label{fig:kf_h128_baseline_RTX5090}
\end{figure*}

\begin{figure*}[h!]
    \centering
    \includegraphics[width=1\textwidth]{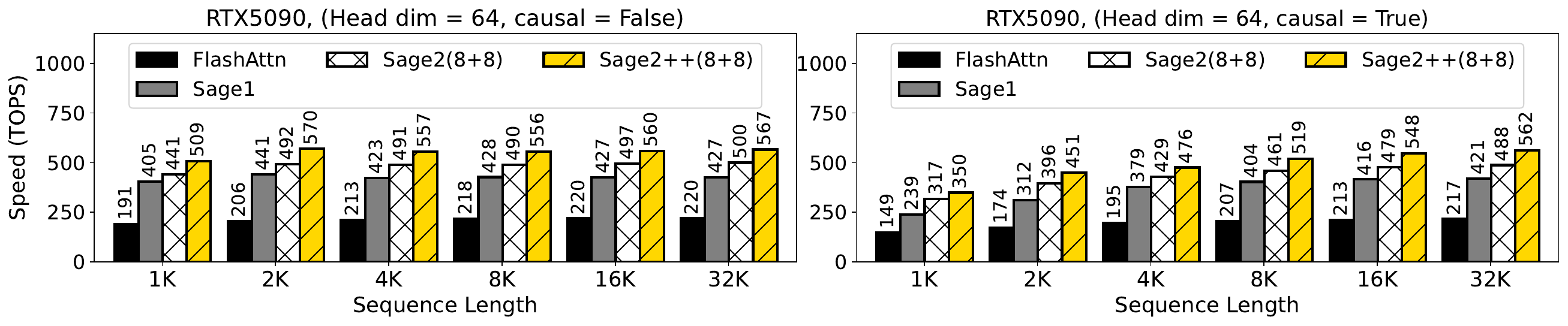}
    \vspace{-1.15em}
    \caption{Speed comparison between \oura and baselines (RTX5090, headdim=64).}
    \vspace{-.5em}
    \label{fig:kf_h64_baseline_RTX5090}
\end{figure*}

\begin{table*}[!t]
    \caption{End-to-end metrics across text, image, and video generation models.
    }
    \label{exp:metrics_loss_t2t}
    \setlength\tabcolsep{21.5pt}
    \begin{center}
    \scalebox{0.95}{\begin{tabular}{p{.8cm}|p{2.2cm}|c|c|c}
    \toprule
    {\bf Model}  & {\bf Attention}  & {\bf WikiText (Ppl.) $\downarrow$}  & {\bf Lambda (Acc.) $\uparrow$}  & {\bf NIAH (Acc.) $\uparrow$}  \\ 
    \hline

    \multirow{3}{*}{\hspace{-1.3em}\llamal} & \hspace{-1.75em}Full-Precision & 6.013  & 0.815  & 0.906   \\  
    & \mbox{\hspace{-1.75em}\oure} & \textbf{6.019} &  \textbf{0.811}  &  \textbf{0.903}  \\ 
    & \mbox{\hspace{-1.75em}\ourea} & \textbf{6.020} &  \textbf{0.813}  &  \textbf{0.901}  \\ 
      \bottomrule
    \end{tabular} }
    \end{center}
\vspace*{-.25em}
    \begin{center}
    \setlength\tabcolsep{13pt}
    \scalebox{0.95}{\begin{tabular}{p{1.45cm}|p{2.8cm}|c|c|c|c|c}
    \toprule
    {\bf Model}  & {\bf Attention}  & {\bf CLIPSIM $\uparrow$}  & {\bf CLIP-T $\uparrow$}  & {\bf VQA-a $\uparrow$}  & {\bf VQA-t $\uparrow$}  & {\bf FScore $\uparrow$} \\ \hline

    \multirow{6}{*}{\hspace{-.75em}\makecell[c]{\cogvideo\\(2B)}} & \hspace{-.5em}Full-Precision  & 0.179 & 0.997 & 74.499 & 74.642 & 4.974 \\
    & \mbox{\hspace{-.5em}\ourf}  & 0.179 & 0.997 & 76.309 & 66.396 & 4.386 \\
    & \mbox{\hspace{-.5em}\oure}  & 0.178 & 0.997 & 74.322 & 74.447 & 4.899 \\
    & \mbox{\hspace{-.5em}\ourfa}  & \textbf{0.179} & \textbf{0.997} & \textbf{74.387} & \textbf{66.568} & \textbf{4.333} \\
    & \mbox{\hspace{-.5em}\ourea}  & \textbf{0.179} & \textbf{0.997} & \textbf{76.309} & \textbf{73.165} & \textbf{4.386} \\
     \hline

    \multirow{6}{*}{\hspace{-.5em}\makecell[c]{\texttt{Hunyuan}\\ \texttt{Video}}} & \hspace{-.5em}Full-Precision & 0.175 & 0.999 & 77.437 & 52.731 & 1.169 \\ 
    & \mbox{\hspace{-.5em}\ourf} & 0.176 & 0.999 & 73.282 & 55.141 & 0.968 \\
    & \mbox{\hspace{-.5em}\oure} & 0.175 & 0.999 & 78.145 & 54.878 & 1.176 \\
    & \mbox{\hspace{-.5em}\ourfa} & \textbf{0.176} & \textbf{0.999} & \textbf{73.282} & \textbf{52.258} & \textbf{0.968} \\
    & \mbox{\hspace{-.5em}\ourea} & \textbf{0.175} & \textbf{0.999} & \textbf{78.569} & \textbf{51.080} & \textbf{1.192} \\
     \hline

    \multirow{6}{*}{\hspace{.5em}\wan} & \hspace{-.5em}Full-Precision & 0.172 & 0.999 & 53.255 & 59.989 & 1.843 \\
    & \mbox{\hspace{-.5em}\ourf} & 0.176 & 0.998 & 29.728 & 38.533 & 0.994 \\
    & \mbox{\hspace{-.5em}\oure} & 0.172 & 0.999 & 49.794 & 55.712 & 1.870 \\
    & \mbox{\hspace{-.5em}\ourfa} & \textbf{0.176} & \textbf{0.998} & \textbf{29.728} & \textbf{38.023} & \textbf{0.994} \\ 
    & \mbox{\hspace{-.5em}\ourea} & \textbf{0.172} & \textbf{0.999} & \textbf{50.876} & \textbf{57.140} & \textbf{1.902} \\
    \bottomrule
    \end{tabular} }
    \end{center}

\vspace*{-.25em}

    \begin{center}
    \setlength\tabcolsep{26.4pt}
    \scalebox{0.95}{\begin{tabular}{p{0.6cm}|p{1.878cm}|c|c|c|c}
    \toprule
    {\bf Model}  & {\bf Attention}  & {\bf FID $\downarrow$}  & {\bf sFID $\downarrow$}  & {\bf CLIP $\uparrow$}  & {\bf IR $\uparrow$}
    \\ \hline

    \multirow{5}{*}{\hspace{-1.2em}\flux} & \hspace{-1.5em}Full-Precision & 165.117 & 147.831 & 31.401 & 0.912 \\
    & \mbox{\hspace{-1.86em}\ourf}  & 164.170 & 147.185 & 31.358 & 0.910 \\
    & \mbox{\hspace{-1.86em}\oure}  & 163.185 & 146.101 & 31.453 & 0.905 \\
    & \mbox{\hspace{-1.86em}\ourfa}  & \textbf{164.170} & \textbf{147.185} & \textbf{31.358} & \textbf{0.910} \\
    & \mbox{\hspace{-1.86em}\ourea}  & \textbf{163.555} & \textbf{146.036} & \textbf{31.445} & \textbf{0.902} \\  \hline
    
    \multirow{5}{*}{\hspace{-2.1em}\texttt{\makecell[c]{Stable-Dif\\fusion3.5}}} & \hspace{-1.5em}Full-Precision & 166.369 & 146.514 & 31.876 & 0.929 \\ 
    & \mbox{\hspace{-1.86em}\ourf}  & 164.610 & 147.350 & 31.912 & 0.914 \\
    & \mbox{\hspace{-1.86em}\oure}  & 164.971 & 148.498 & 31.964 & 0.931 \\
    & \mbox{\hspace{-1.86em}\ourfa}  & \textbf{164.610} & \textbf{147.350} & \textbf{31.912} & \textbf{0.914} \\
    & \mbox{\hspace{-1.86em}\ourea}  & \textbf{165.842} & \textbf{146.465} & \textbf{31.968} & \textbf{0.929} \\
\bottomrule
    \end{tabular} }
    \end{center}
\vspace{-.6em}
\end{table*}

\begin{figure*}[h!]
    \centering
    \includegraphics[width=1\textwidth]{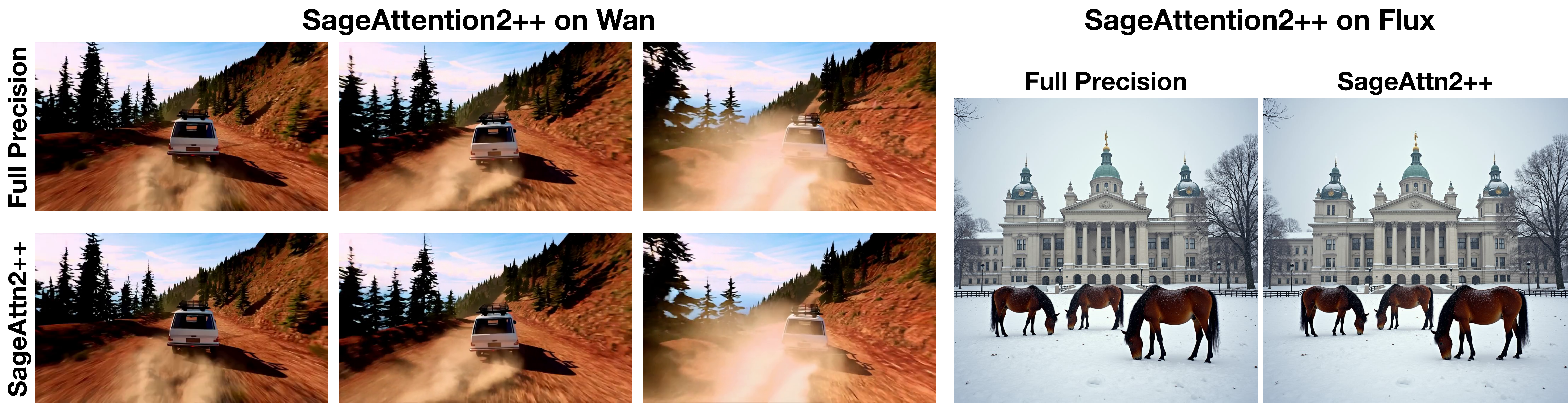}
    \vspace{-1em}
    \caption{A visible example of using \oura.}
    \vspace{-1em}
    \label{fig:visible_example}
\end{figure*}

\begin{figure*}[h!]
    \centering
    \includegraphics[width=0.99\textwidth]{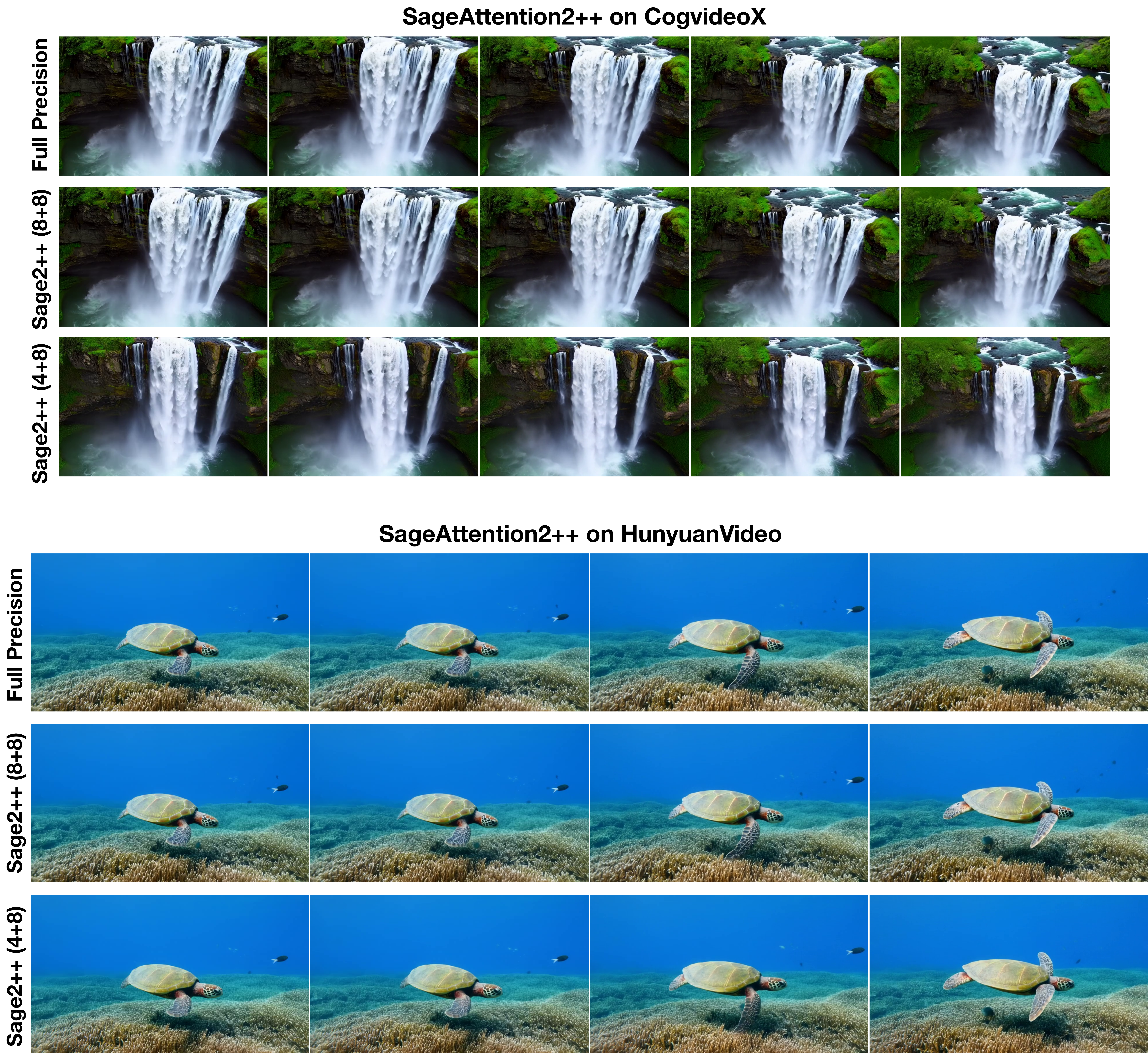}
    \vspace{-.5em}
    \caption{Visible examples of using \oura on video generation.}
    \vspace{-.5em}
    \label{fig:video_example}
\end{figure*}

\section{Experiment}
\textbf{Main result.} \oura achieves up to \textbf{3.9}$\times$ speedup over FlashAttention2 while consistently outperforming both SageAttention and SageAttention2 in computational efficiency. Importantly, these performance gains are achieved with negligible impact on end-to-end metrics across diverse model architectures.

\subsection{Setup} \label{sec:exp}
\noindent\textbf{Models and attentions.} 
We evaluate \oura across diverse representative models spanning language, image, and video generation: \llamal(8B)~\cite{llama31model} for text2text, \cogvideo(2B), \hyvideo~\cite{kong2024hunyuanvideo}, and \wan~\cite{wan2025} for text2video, and \flux(schnell)~\cite{flux} and \sd(turbo)~\cite{stable_diffusion_3_5} for text2image. We compare our method with FlashAttention2~\citep{dao2023flashattention}, SageAttention~\citep{2024sageattention}, and SageAttention2~\citep{zhang2024sageattention2}. Please \textbf{note} that FlashAttention3 can only run on Hopper GPUs, so FlashAttention2 is already the fastest version for {RTX5090} and {RTX4090}. Following SageAttention2's approach, we implement two \oura variants: \ourea (INT8 for $Q,K$) and \ourfa (INT4 for $Q,K$), both using FP8 in E4M3 for $\widetilde P,V$.

\noindent \noindent\textbf{Datasets and metrics.} Detailed dataset and metric information appears in Appendix~\ref{sec:exp_dataset_metrics}.

\noindent\textbf{Implementation.} We implement \oura using CUDA.

\subsection{Speed of Kernels}

\noindent\textbf{Kernel Speed.} We benchmark the speed of \oura against baselines using configurations with headdim=64 and headdim=128, both with and without a Causal Mask~\cite{vaswani2017attention}. Specifically, Fig.~\ref{fig:kf_h128_baseline_RTX4090} shows the speed across varying sequence lengths on RTX4090, indicating that \ourfa and \ourea are approximately 3.9x and 3.0x faster than FlashAttention2, respectively.
Fig.~\ref{fig:kf_h64_baseline_RTX4090}, ~\ref{fig:kf_h128_baseline_RTX5090} and~\ref{fig:kf_h64_baseline_RTX5090} show more kernel speed comparison on RTX4090 and RTX5090 GPUs.

\subsection{End-to-end Performance}

\noindent\textbf{Metrics loss.} We evaluate end-to-end model performance using \oura against baseline methods.
Detailed evaluation results are presented in Table~\ref{exp:metrics_loss_t2t}.
The results indicate that \ourea and \ourfa match the end-to-end metrics of SageAttention2. Specifically, \ourea incurs almost no metrics loss across various models and \ourfa brings a little metrics loss.

\noindent\textbf{Visible image and video examples.} Fig.\ref{fig:visible_example}, ~\ref{fig:image_example}, and ~\ref{fig:video_example} show some visible comparison examples.

%% file: src/Conclusion.tex
\section{Conclusion}
We introduce \oura to further accelerate SageAttention2. We propose to utilize the faster instruction of FP8 Matmul accumulated in FP16 for the matrix multiplication of $PV$. 
Experiments show that \oura achieves a \textbf{3.9}$\times$ speedup over FlashAttention while maintaining the same attention accuracy as SageAttention2. This means \oura can accelerate various models, including those for language, image, and video generation, with negligible end-to-end metrics loss.

%% file: src/Appendix.tex
\section{Appendix}
\subsection{Visible Comparison Examples}  \label{sec:appedix_visible_example}

\begin{figure*}[h!]
    \centering
    \includegraphics[width=0.81\textwidth]{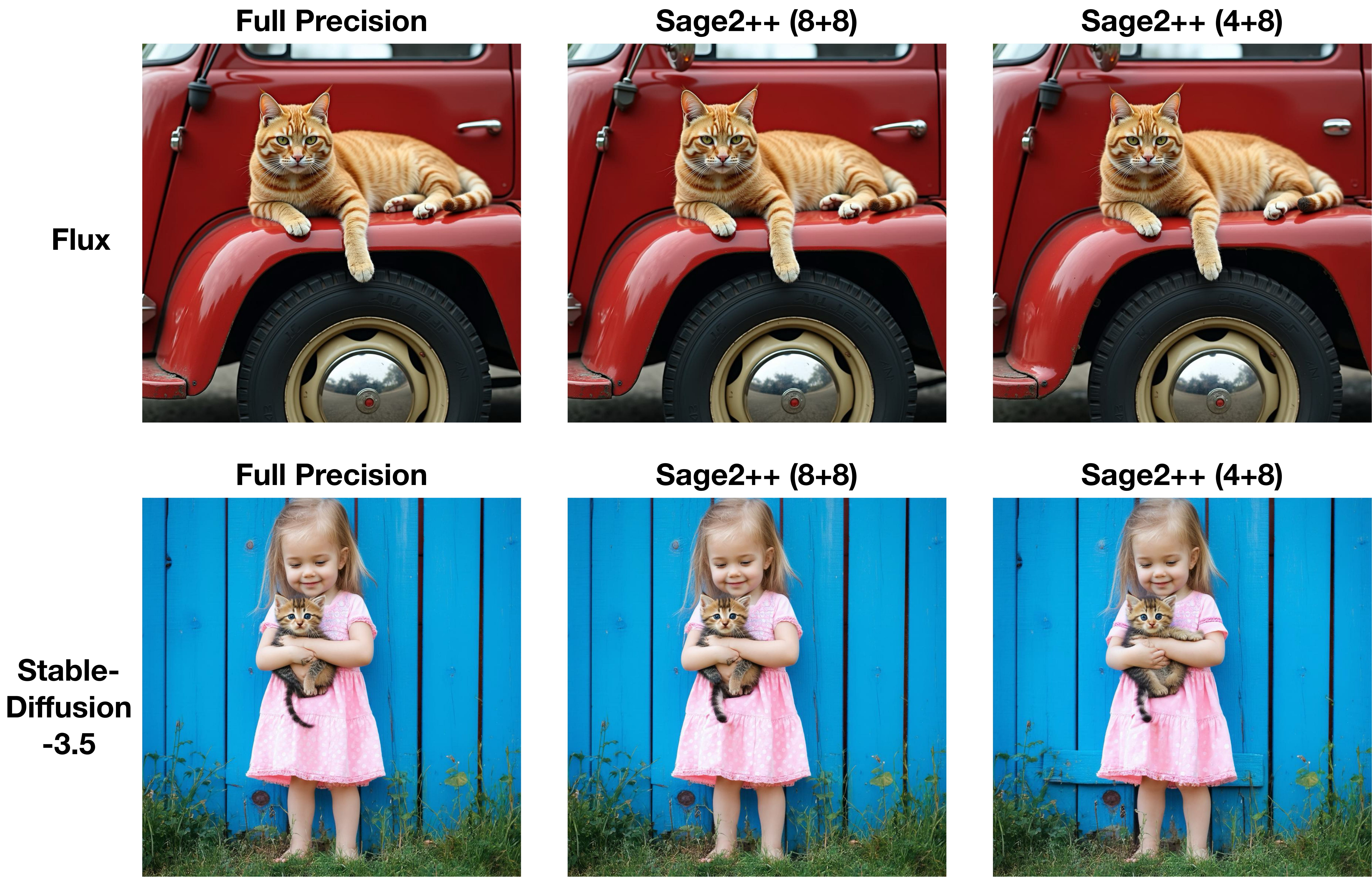}
    \vspace{-1em}
    \caption{Visible examples of using \oura on image generation.}
    \vspace{-.5em}
    \label{fig:image_example}
\end{figure*}

\subsection{Datasets and Metrics in Experiments}  \label{sec:exp_dataset_metrics}

\noindent \noindent\textbf{Datasets.}   
Text-to-text models are evaluated on: WikiText~\cite{merity2022pointer} to assess the model's prediction confidence, LAMBADA~\cite{paperno2016lambada} for contextual understanding, and Needle-in-A-Haystack (NIAH) task~\cite{LLMTest_NeedleInAHaystack}.
Text-to-video models are evaluated using the open-sora~\cite{opensora} prompt sets.
Text-to-image models are assessed on COCO annotations~\cite{lin2014microsoft}. 

\noindent \noindent\textbf{End-to-end metrics.}   
For text-to-text models, we use perplexity (Ppl.)~\cite{jelinek1977perplexity} for WikiText, accuracy (Acc.) for LAMBADA and NIAH.
For text-to-video models, following~\citet{zhao2024viditq}, we evaluate the quality of generated videos on five metrics: CLIPSIM and CLIP-Temp (CLIP-T)~\cite{liu2024evalcrafter} to measure the text-video alignment; VQA-a and VQA-t to assess the video aesthetic and technical quality, respectively; and Flow-score (FScore) for temporal consistency~\cite{wu2023exploring}. 
For text-to-image models, generated images are compared with the images in three aspects: FID~\cite{heusel2017gans} and sFID~\cite{salimans2016improved} for fidelity evaluation, \textit{Clipscore} (CLIP)~\cite{hessel2021clipscore} for text-image alignment, and \textit{ImageReward} (IR)~\cite{xu2024imagereward} for human preference.

\noindent\textbf{Accuracy metrics.} We use three metrics to assess the accuracy of quantized attention output $O'$ compared to attention output in full-precision $O$. First, we flatten $O'$ and $O$ into vectors in the shape of $1\times n$. Then, Cosine similarity: $CosSim=\sum OO' / \smash{\sqrt{\sum O^2}} \smash{\sqrt{\sum O'^2}}$, Relative L1 distance: $L1=\sum |O - O'| / \sum |O|$, Root mean square error: $RMSE=\sqrt{(1/n) \sum (O - O')^2}$.